\documentclass[lettersize,journal]{IEEEtran}

\usepackage{lineno}
\usepackage[percent]{overpic}
\usepackage{amsmath,amsfonts}
\usepackage{algorithmic}
\usepackage{array}
\usepackage[caption=false,font=normalsize,labelfont=sf,textfont=sf]{subfig}
\usepackage{textcomp}
\usepackage{stfloats}
\usepackage{url}
\usepackage{verbatim}
\usepackage{graphicx}
\usepackage{xcolor}
\usepackage{cite}
\usepackage{booktabs} 
\usepackage{multirow}
\usepackage[ruled,vlined]{algorithm2e}
\SetAlFnt{\small}

\newif \ifrevise
\revisefalse
\ifrevise
    \usepackage{lineno}
    \newcommand{\modify}[1]{\textcolor{blue}{#1}}
  \else
    \newcommand{\modify}[1]{\textcolor{black}{#1}}
\fi

\definecolor{forestgreen}{HTML}{009B55} 

\definecolor{commentcolor}{RGB}{110,154,155}   
\definecolor{keycolor}{rgb}{0.858, 0.188, 0.478} 

\def\BibTeX{{\rm B\kern-.05em{\sc i\kern-.025em b}\kern-.08em
    T\kern-.1667em\lower.7ex\hbox{E}\kern-.125emX}}

\begin{document}
\ifrevise
	\pagewiselinenumbers
	\switchlinenumbers
\fi

\title{A Clustering-guided Contrastive Fusion for Multi-view Representation Learning}

\author{Guanzhou Ke,
        Guoqing Chao,
        Xiaoli Wang, 
        Chenyang Xu,
        Yongqi Zhu,
        and Yang Yu
\thanks{G. Ke, Y. Zhu, and Y. Yu are with the Institute of Data Science and Intelligent Decision Support, Beijing Jiaotong University, Beijing 100080, China. E-mail: \{guanzhouk, 21120632, yangy1\}@bjtu.edu.cn}
\thanks{G. Chao is with the School of Computer Science and Technology, Harbin Institute of Technology, Weihai 264209, China, E-mail: guoqingchao@hit.edu.cn}
\thanks{X. Wang is with the School of Computer Science and Engineering, Nanjing University of Science And Technology, Nanjing 210000, China. E-mail: xiaoliwang@njust.edu.cn}
\thanks{C. Xu is with the Faculty of Intelligent Manufacturing, Wuyi University, Jiangmen 529000, China. E-mail: chyond.xu@gmail.com}
\thanks{(Corresponding author: Yang Yu.)}
}

\markboth{Journal of \LaTeX\ Class Files,~Vol.~14, No.~8, August~2021}%
{Guanzhou Ke \MakeLowercase{\textit{et al.}}: A Clustering-guided Contrastive Fusion for Multi-view Representation Learning}



\maketitle

\begin{abstract}
Multi-view representation learning aims to extract comprehensive information from multiple sources. It has achieved significant success in applications such as video understanding and 3D rendering. However, how to improve the robustness and generalization of multi-view representations from unsupervised and incomplete scenarios remains an open question in this field. In this study, we discovered a positive correlation between the semantic distance of multi-view representations and the tolerance for data corruption. Moreover, we found that the information ratio of consistency and complementarity significantly impacts the performance of discriminative and generative tasks related to multi-view representations. Based on these observations, we propose an end-to-end CLustering-guided cOntrastiVE fusioN (CLOVEN) method, which enhances the robustness and generalization of multi-view representations simultaneously. To balance consistency and complementarity, we design an asymmetric contrastive fusion module. The module first combines all view-specific representations into a comprehensive representation through a scaling fusion layer. Then, the information of the comprehensive representation and view-specific representations is aligned via contrastive learning loss function, resulting in a view-common representation that includes both consistent and complementary information. We prevent the module from learning suboptimal solutions by not allowing information alignment between view-specific representations. We design a clustering-guided module that encourages the aggregation of semantically similar views. This action reduces the semantic distance of the view-common representation. We quantitatively and qualitatively evaluate CLOVEN on five datasets, demonstrating its superiority over 13 other competitive multi-view learning methods in terms of clustering and classification performance. In the data-corrupted scenario, our proposed method resists noise interference better than competitors. Additionally, the visualization demonstrates that CLOVEN succeeds in preserving the intrinsic structure of view-specific representations and improves the compactness of view-common representations. Our code can be found at \url{https://github.com/guanzhou-ke/cloven}.

\end{abstract}


\section{Introduction}

\IEEEPARstart{T}{he} goal of multi-view representation learning (MvRL)\cite{wang2015deep} is to learn an encoding that maps either homogeneous (such as CCTV images from different angles) or heterogeneous (such as images and audio in videos) multi-source data into a shared representation space, also called a view-common representation.
\begin{figure}[t]
\centering
\includegraphics[width=3.5in]{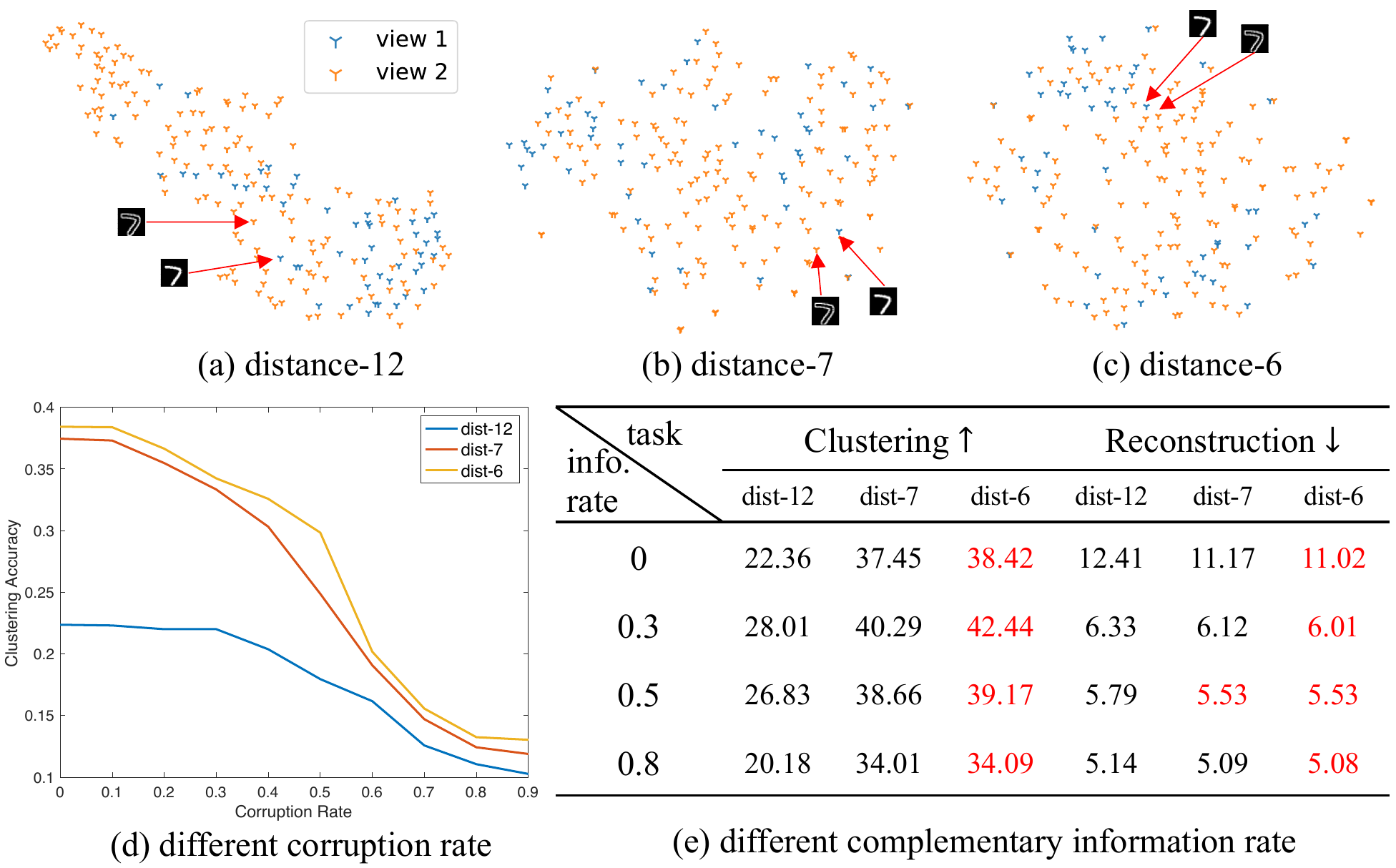}
\caption{
\modify{We conducted controlled experiments on the Edge-MNIST dataset~\cite{liu2016coupled} to illustrate our method's key observations and basic idea. Three models with different average intra-cluster semantic distances (dist-12, -7, and -6) were investigated under varying data corruption rates and tasks. Figures (a)-(c) depict two views' representations of the digit ``7'' with 100 samples under different semantic distance models. Figure (d) presents the clustering accuracy of the three models under different data corruption degrees, assessed using k-means. It is evident that models with tighter average semantic distances exhibit higher tolerance to data corruption. We utilized $\beta$-VAE~\cite{higgins2017beta} to decouple consistency and complementarity, with a fixed concatenating representation dimension of 10. By controlling the ratio of complementary information, we observed changes in the clustering and reconstruction performance of the three models. Clustering accuracy and mean squared error were used to evaluate clustering and reconstruction, respectively. When the information rate was set to 0.3, indicating a consistent representation dimension is $7$ and a complementary representation dimension is $3$, a balance point was reached in terms of clustering and reconstruction performance for the three models. Notably, models with tighter average semantic distances achieved higher performance. As the ratio of complementary information increased, the reconstruction task performance improved while the clustering task performance declined. This observation underscores the significance of balancing consistency and complementarity in enhancing the generalization of multi-view representations.}} 
\label{pic:observations}
\end{figure}
MvRL has achieved remarkable success in various downstream tasks over the past few decades\cite{LiYZ19}, such as classification\cite{kan2016multi, you2019multi, zhou2019multi, jia2020semi, wang2022mmatch} and video action recognition\cite{wang2020learning, zheng2021collaborative, wang2019generative, tran2022effective}. These accomplishments are attributed to the community's accumulation of a large amount of annotated data and expensive computational resources. However, improving the robustness and generalization of multi-view representations in unsupervised and incomplete scenarios remains a challenge in this field. Robustness refers to multi-view applications' ability to train and test in incomplete scenarios, such as data corruption. Generalization implies that the extracted representations should adapt to different types (discriminative or generative) of downstream tasks in a cost-effective manner.

A number of successful techniques have been developed to address the task of learning effective multi-view representations from large-scale unlabeled data. These include unsupervised methods such as canonical correlation analysis (CCA) \cite{an2014face, xing2016complete, jiang2016srlsp}, subspace-based methods \cite{gao2015multi, zhang2020tensorized}, and deep learning methods \cite{kan2016multi, abavisani2018deep, zhang2019ae2, andrew2013deep, wang2015deep}. In order to enhance the robustness of representations, previous research has focused on training models to learn view-invariant representations using adversarial learning \cite{zhou2020eamc, li2019damc} or data augmentation techniques \cite{tain2020cmc, Federici0FKA20}. Conversely, other works have focused on maximizing inter-view consistency or incorporating auxiliary losses, such as clustering \cite{lin2021completer, ke2022efficient}. Although the core idea behind these approaches is to enhance the robustness of multi-view representations by maximizing consistency across views, this objective can be task-dependent, resulting in reduced generalization performance of learned representations. While consistency is important for clustering \cite{abavisani2018deep, trosten2021reconsidering, huang2021joint} or classification tasks \cite{wang2022mmatch, jia2020semi}, the complementary information contained within multi-views is critical for specific tasks like cross-view information retrieval \cite{yan2020deep, liu2021hierarchical, hu2020joint} and multi-view synthesis \cite{chen2019point, riegler2021stable}. Thus, improving the generalization of multi-view representations is key for expanding the range of applications to which it can be applied. Various techniques have been developed to achieve this end, including fusion-based methods \cite{ke2022efficient, ke2021conan, wang2020deep, xia2022incomplete} or autoencoder-based paradigms \cite{zhang2019ae2, abavisani2018deep, lin2021completer, xu2021multi}. While these approaches effectively improve the generalization of multi-view representations, the integration of complementary information can also reduce their robustness.

Achieving both robustness and generalization of multi-view representations has been a challenging problem in previous research. As a result, we explore a unified framework that balances both aspects by investigating the average semantic distance between multi-view representations, as presented in \figurename{\ref{pic:observations}}. Conducting controlled experiments\footnote{See our release code for the experiment details.} on the Edge-MNIST~\cite{liu2016coupled} dataset, we trained models with different average semantic distances between intra-cluster representations. As shown in \figurename{\ref{pic:observations}} (d) and (e), we found that models with tighter semantic distances between their representations demonstrated better tolerance for data corruption. We further studied representation performance variations in clustering and reconstruction tasks under different ratios of multi-view consistent and complementary information. Our observations show that compact semantic distance of representations improves their generalization. Additionally, these findings revealed that balancing consistency and complementarity is critical to improving generalization.

Drawing on the aforementioned observations, we believe that it is possible to simultaneously enhance both the robustness and generalization of multi-view representations by reducing the intra-view semantic distance. A natural idea is to utilize clustering as a guiding principle. To ensure the generalization of representations, we obtain semantically compact and expressive view-common representations by fusing all view-specific representations. Therefore, we introduce a multi-view fusion framework, called the CLustering-guided cOntrastiVE fusioN (CLOVEN). First, we use deep fusion networks to extract the robust and expressive common representation from view-specific representations. We describe two versions of deep fusion networks: vanilla MLPs and residual-based MLPs (explained in Section \ref{sec:deep-fusion-networks}). Second, we introduce a clustering-guided mechanism that helps the fusion networks learn a compact view-common representation. Then, we propose our novel contrastive strategy, called the asymmetrical contrastive strategy, which only allows view-specific representations to interact in the view-common space to prevent suboptimal solutions. Unlike previous works \cite{lin2021completer, tain2020cmc, trosten2021reconsidering}, this approach effectively overcomes the problem of suboptimal solutions. Our approach demonstrates good generalization on five clustering and classification datasets, as shown by experimental results. Additionally, we evaluate CLOVEN's robustness in two different incomplete view scenarios and show that it can effectively overcome data corruption. We found through visualization analysis that the asymmetrical contrastive strategy can preserve the intrinsic structure of each view-specific representation and that the clustering-guided mechanism can enhance the compactness of the view-common representation. The contributions of this paper are summarized as follows:

\begin{enumerate}
\item We discovered a positive correlation between the robustness and generalization of multi-view representations and the intra-view semantic distance. Therefore, we propose a clustering-guided contrastive fusion method that improves the simultaneous robustness and generalization. Our method integrates the contrastive fusion module and the clustering-guided method into a unified framework to learn view-common representation in an end-to-end manner.
\item \modify{We conducted experiments on five real-world datasets to demonstrate the effectiveness of our method, CLOVEN, in terms of clustering and classification performance. Visualization results confirmed the successful preservation of intrinsic structure in view-specific representation through an asymmetrical contrastive strategy, while the clustering guidance mechanism improved the compactness of the view-common representation. Additionally, CLOVEN's extracted view-common representation not only reduces computational costs in downstream tasks but also enhances overall performance.}
\item We evaluated CLOVEN's robustness in two incomplete view scenarios and observed that its performance remains superior to most algorithms, even with a corruption rate above 0.4. Furthermore, our findings highlight the effectiveness of the pre-trained view-specific encoder in resisting corruption interference.
\end{enumerate}

\section{Related Works}\label{sec:related-works}

\subsection{Multi-view Representation Learning}
In general, multi-view representation learning can be roughly classified into two categories: statistic-based methods \cite{peng2019comic, zheng2021collaborative, you2019multi, wen2019unified, zhang2018binary, liu2020efficient, liu2013multi} and deep learning-based methods \cite{kan2016multi, abavisani2018deep, zhang2019ae2, andrew2013deep, wang2015deep}. Statistic-based methods include  multi-kernel\cite{liu2019multiple, liu2015multiple, zhang2022sample}, subspace\cite{gao2015multi, zhang2020tensorized, an2014face}, and CCA-based approaches\cite{xing2016complete, xing2016complete, jiang2016srlsp, peng2019comic, wen2019unified, zhang2018binary, liu2020efficient, liu2013multi}. These methods are not suitable for high-dimensional and complex data scenarios. Zheng et al. \cite{zheng2021collaborative} introduced a low-rank tensor constraint to enhance the compactness of view-specific and view-common representations by exploring higher-order cross-view relationships. To address the problem of data corruption, Wong et al. \cite{wong2019clustering} analyzed three types of noise structures, namely, symmetric, column-wise, and row-wise, and suggested minimizing dissonance over full views to extract consistency. Our method differs from these approaches by considering the high-level distance relationship between multi-view representations' semantics. Consequently, our approach has a broader coverage of scenarios where data corruption occurs.

In the past two decades, a plethora of deep learning-based approaches have been developed to tackle the challenges of large-scale data. For instance, CCA-based methods \cite{andrew2013deep, wang2015deep} as well as subspace-based methods \cite{abavisani2018deep, li2019damc}, etc. have been proposed. Zheng et al. \cite{zheng2022graph} proposed a graph-guided representation learning technique that captures the higher-order structure of view-common representations by exploring relationships between view-specific graphs.Wang et al.\cite{wang2022mmatch} introduced a semi-supervised method to enhance the discriminative capability of the model for weak discriminative views by enforcing constraints of consistency between local view structures and global view structures, while leveraging global structural information for appropriate pseudo-label inference. Zhu et al. \cite{zhu2022latent} proposed a graph-based incomplete view method by introducing the neighborhood constraint and view-existence constraint to create a heterogeneous graph. The heterogeneous graph distinguishes the relationship between the crucial graph nodes and their neighbors, thus capturing the complex relationship between views. In an unsupervised environment, most of these approaches rely on generative models such as autoencoder-based methods \cite{andrew2013deep, wang2015deep, lin2021completer, ke2022efficient, zhang2019ae2, zhang2020deep} and generative adversarial networks-based methods (GANs) \cite{li2019damc, zhou2020eamc} to acquire latent representations from data. Although these models can extract valuable information from data to improve downstream tasks, they have some disadvantages, such as a lengthy training process \cite{li2019damc, ke2022efficient} and the need to retain extensive information to reconstruct the original data. To tackle these challenges, researchers have started to explore contrastive learning-based approaches such as \cite{lin2021completer, 10031085, ke2021conan, trosten2021reconsidering, xu2022multi}. Discriminative modelsare inclined to acquire conceptual information from multiple views data, in contrast to generative models. Differently from past methodologies, our proposed approach integrates fusion and alignment into a unified framework using contrastive learning, and not only to align view-specific representations. In contrast to the prior works \cite{lin2021completer, trosten2021reconsidering, xu2022multi}, we use the asymmetrical contrastive strategy to align every view-specific representation through the view-common representation obtained after the fusion, rather than directly aligning on the view-specific representation. 

\begin{table}[ht]
\caption{Summary of the key notations used in the paper\label{tab:key-notations}}
\centering
\scalebox{0.9}{\begin{tabular}{cc}
\specialrule{0.05em}{3pt}{3pt}
Notations & Explanations\\
\specialrule{0.05em}{3pt}{3pt}
$N, V$ & the number of samples and views \\
$L$ & the layers of fusion module \\
$k$ & the number of clusters \\
$d(\cdot)$ & the dimension measure function\\
$\{\mathbf{X}^{(i)}\}^V_{i=1}$ & multi-view data of $V$ views with $ \mathbf{X}^{(i)} \in \mathbb{R}^{n \times d(\mathbf{X}^{(i)})}$ \\
$\{\mathbf{H}^{(i)}\}^V_{i=1}$ & view-specific representation of $V$ views with $ \mathbf{H}^{(i)} \in \mathbb{R}^{n \times d(\mathbf{H}^{(i)})}$\\
$\mathbf{Z}$ & view-common representation with $ \mathbf{Z} \in \mathbb{R}^{n \times d(\mathbf{Z})}$\\
$E_i(\cdot)$ & the $i$-th view-specific encoder. \\
$f(\cdot)$ & the fusion module. \\
$g(\cdot)$ & the clustering head. \\
$M(\cdot)$ & a multilayer perceptron.  \\
$[\cdot]$ & the concatenating operation. \\
\specialrule{0.05em}{3pt}{3pt}
\end{tabular}}
\end{table}

\subsection{Contrastive Learning}
Contrastive learning is a metric learning technique that aims to reduce the distance among similar samples while increasing the distance among dissimilar samples in a contrastive space. In single-view scenarios, positive and negative sample pairs are constructed using data augmentation techniques \cite{chen2020simclr}. Different augmentations of the same object are considered positive samples, while augmentations of different objects are negative samples. In computer vision research, numerous studies such as \cite{chen2020simclr, he2020momentum, chen2021exploring, grill2020bootstrap} have shown the potential of contrastive learning to obtain latent representations from data. Tain et al. \cite{tain2020cmc} extend contrastive learning from single-view to multi-view scenarios. Typically, multi-view contrastive learning assumes the existence of a latent object and that the data collected from different perspectives (or views) is an augmentation of that object. Recently, numerous multi-view approaches based on contrastive learning such as \cite{lin2021completer, lin2022dual, ke2021conan, tain2020cmc, trosten2021reconsidering, xu2022multi} have achieved remarkable success. Directly aligning the distribution on view-specific representations can cause the destruction of the intrinsic structure of the view, as noted in \cite{trosten2021reconsidering}. To mitigate this problem, the method proposed in \cite{trosten2021reconsidering} is to use weighted-sum fusion to determine the importance of each view based on contrastive learning. Unlike previous approaches, we use an indirect approach based on the alignment of fused representations and bypass direct interaction between view-specific representations. In our context, if we allow for alignment among view-specific representations, then the intrinsic structure of each view could be disrupted, increasing the risk of model collapse since the view-common representation is fused from all view-specific representations. Li et al. \cite{li2021contrastive} proposed a single-view contrastive clustering method which obtains semantically rich representations by using pseudo-label consistency and augmenting transformation invariance. To balance consistency and complementarity between views, we extend this approach to the multi-view scenario. Nevertheless, our approach differs from \cite{li2021contrastive} in that we use the asymmetric alignment strategy that only allows for alignment between view-specific representations and comprehensive representations, rather than alignment among views. \cite{deng2023heterogeneous, deng2023strongly} have employed strong augmentation strategies to learn the consistency between heterogeneous views. These works also employ instance-level and category-level contrastive losses, while our proposed method is more focused on using clustering as a guidance mechanism to enhance multi-view robustness by enhancing the compactness between representations.

\subsection{Multi-view Fusion}

The objective of multi-view fusion is to integrate multiple view-specific representations into a unified representation as known as view-common representation. Fusion techniques can be approximately classified into two types: shallow fusion and deep fusion. For example, concatenating \cite{li2019damc, lin2021completer} and weighted-sum \cite{zhou2020eamc, trosten2021reconsidering} are two common approaches for shallow fusion. The former combines all view-specific representations into a high-dimensional representation form with the most information-rich representation, may cause expensive computational burden in downstream tasks. The latter extracts the view-common representation if the mapped representation space can be added linearly and has the same dimensionality as the view-specific representation, but it may lead to suboptimal solutions otherwise. Unlike shallow fusion, deep fusion frequently adopts either graphic-based or neural-based techniques, as found in \cite{peng2019comic, abavisani2018deep, huang2021deep} and \cite{ngiam2011multimodal, ke2022efficient, tu2021deep}, respectively. Deep fusion enhances the expressiveness of the view-common representation by extracting hierarchical semantic information. Deep fusion is more prevalent than graphic-based fusion as it does not need a preconception about the data distribution. \cite{wang2020deep} proposed an adaptive fusion method, where they scaled the representation of each view, and then completed the channel-level fusion of each view with Batch Normalization. This study showed that scaling representations can effectively balance view consistency and complementary. Unlike \cite{wang2020deep}, we do not utilize Batch Normalization for aligning the feature scaling. Instead, we use the feature dimension to scale the representations. The advantage of CLOVEN is that we can adapt to various batch sizes and application scenarios. In \cite{xia2022incomplete}, the authors assign fusion weights to each view by learning the similarity between the consistency and complementarity of views. Finally, the multi-view representations are fused by means of a weight-sum method that considers the information overlap and content disparity among views. The method considers the overlap between consistency and complementarity as well as the difference in view information. Our method, however, does not rely on explicit weight assignment but rather fuses representations by minimizing the distance between semantically similar ones. In this paper, We commence with the fundamental form of neural network fusion and further expand it to the residual-based fusion method that facilitates feature interaction. Unlike previous methods, we design the feature scaling module in the form of an information bottleneck \cite{Federici0FKA20} to prevent obtaining suboptimal solutions and to enhance the generalization and robustness of the view-common representation.

\section{Method}\label{sec:method}

\begin{figure*}[t]
\centering
\includegraphics[width=6.8in]{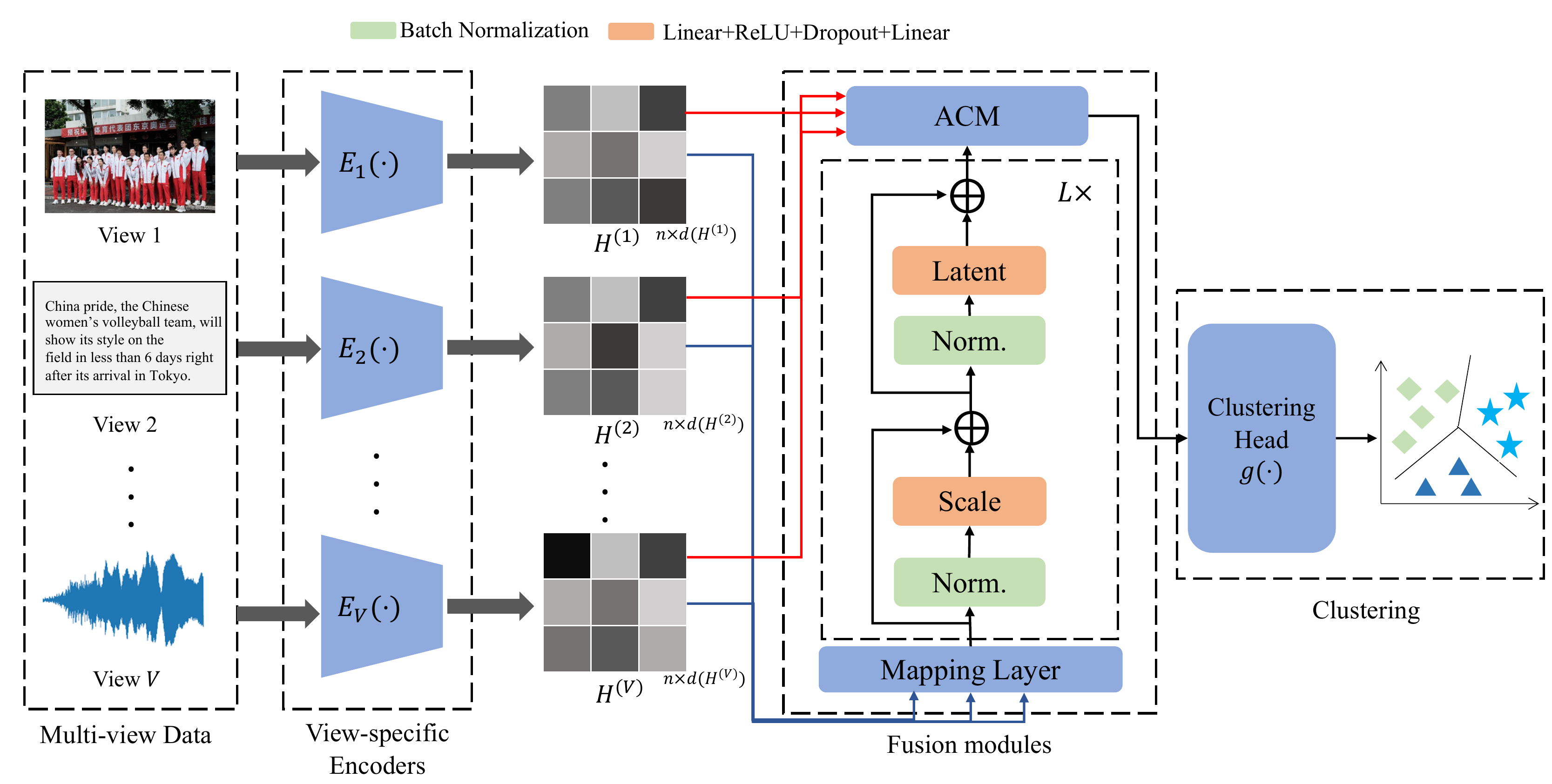}
\caption{Illustration of the workflow of CLOVEN. It consist of $V$ view-specific encoder networks $E_i(\cdot)$ to obtain view-specific representations $\{\mathbf{H}^{(i)}\}^{V}_{i=1}$ from multi-view data $\{\mathbf{X}^{(i)}\}^{V}_{i=1}$, a fusion module $f(\cdot)$ to obtain the view-common representations $\mathbf{Z}$ via fusion the view-specific representations. In ACM, we treat $\mathbf{Z}$ and $\{\mathbf{H}^{(i)}\}^{V}_{i=1}$ as input, and we employed an asymmetrical contrastive strategy to maximize the mutual information between $\mathbf{Z}$ and each $\mathbf{H}^{(i)}$. Finally, we apply an off-the-shelf clustering method on $\mathbf{Z}$, extracting the view consistent information.}
\label{pic:framework}
\end{figure*}

\subsection{Problem Formulation}\label{sec:problem-formulation}
The objective of CLOVEN is to obtain a robust and diverse set of view-common representation, denoted as $\mathbf{Z}$, from $N$ data points consisting of $V$ views $\{\mathbf{X}^{(1)}, \mathbf{X}^{(2)}, ..., \mathbf{X}^{(V)}\}$. Here, $\mathbf{X}^{(i)} \in \mathbb{R}^{N \times d(\mathbf{X}^{(i)})}$ indicates the samples of dimension $d(\mathbf{X}^{(i)})$ for the $i^{th}$ view. The framework of CLOVEN is presented in Fig. \ref{pic:framework}, and the key notations employed in this paper are listed in Table \ref{tab:key-notations}. To start with, we first use $V$ view-specific encoders, such as fully-connected networks or convolutional networks, to extract the corresponding view-specific representation $\mathbf{H}$. The view-specific representation $\mathbf{H}$ for the $i^{th}$ view is represented as $\mathbf{H}^{(i)} \in \mathbb{R}^{N \times d(\mathbf{H}^{(i)})}$,
\begin{equation}
\label{eq:extract_vsr}
 \mathbf{H}^{(i)} = E_i(\mathbf{X}^{(i)}; \mathbf{W}_E^{(i)}); \quad \text{where} \quad i = 1, 2, \cdots , V
\end{equation}
$\mathbf{W}_E^{(i)}$ denotes the weights of the $i^{th}$ view-specific encoder $E_i$. Note that the view-specific encoder can be any type of neural network, such as fully-connected networks or convolutional networks. A mapping layer, which is a fully-connected network, is then used to project all view-specific representations $\mathbf{H}$ to low-dimensional space. By feeding these low-dimensional codes to the stackable fusion blocks, we obtain the view-common representation $\mathbf{Z}$ after dimensional scaling and latent space encoding, as shown in (\ref{eq:fusion}).
\begin{equation}
\label{eq:fusion}
 \mathbf{Z} = f([\mathbf{H}^{(1)}, \mathbf{H}^{(2)}, \cdots, \mathbf{H}^{(V)}]; \mathbf{W}_f)
\end{equation}
Here, $\mathbf{Z} \in \mathbb{R}^{N \times d(\mathbf{Z})}$, and the details of the fusion process are explained in Section \ref{sec:deep-fusion-networks}. To improve the robustness, an asymmetrical contrastive module (ACM) is designed for contrastive fusion. Additionally, clustering is used as a guiding mechanism to enhance the view consistency in $\mathbf{Z}$. In the following subsections, we delve into the specificities of these modules.

\subsection{Deep Fusion Module}\label{sec:deep-fusion-networks}
The aim of fusion is to merge view-specific representations into a cohesive representation space called a view-common representation. In contrast to shallow fusion methodologies such as the concatenate  \cite{lin2021completer, li2019damc} and weight-sum \cite{zhou2020eamc, trosten2021reconsidering} approaches, our goal is to develop a deep fusion paradigm that extracts valuable features progressively. This approach is expected to deliver view-common representations that are semantically rich, space-saving, and more expressive. A basic idea to achieve deep fusion is to leverage stacked fully-connected networks, or vanilla MLP, to map all view-specific representations to a low-dimensional representation, as shown in (\ref{eq:vanilla-fusion}). This method is advantageous due to its simplicity and compatibility with current multi-view methodologies, with previous studies \cite{abavisani2018deep, rastegar2016mdl, ke2021conan} classified into such a category.
\begin{equation}
\label{eq:vanilla-fusion}
f_{vanilla} = M([\mathbf{H}^{(1)}, \mathbf{H}^{(2)}, \cdots, \mathbf{H}^{(V)}])
\end{equation}

Previous research by He et al.\cite{he2016deep} indicated that as the network depth increases, the network might converge to a suboptimal solution. Therefore, we introduced the residual block (RB) or skip connection, as depicted in (\ref{eq:residual-block}).
\begin{equation}
\label{eq:residual-block}
RB(z) = M(norm(z)) + z
\end{equation}
where $norm(\cdot)$ denotes the batch normalization, $z$ is the intermediate embedding of the view-common representation $\mathbf{Z}$. To obtain the intermediate embedding from all view-specific representations, we leverage a mapping layer composed of a single MLP to map the high-dimensional view-specific representation to the intermediate embedding $z$, i.e., $z = M([\mathbf{H}^{(1)}, \mathbf{H}^{(2)}, \cdots, \mathbf{H}^{(V)}])$. To enhance the expressiveness of view-common representation $\mathbf{Z}$ as proposed in \cite{he2016deep,vaswani2017attention}, we introduce two functional modules: ScaleBlock(SB) and LatentBlock(LB). SB doubles the dimension of the intermediate embedding $z$, maps it back to the original dimension, while LB reduces the dimension of intermediate embedding $z$ to half of the original dimension and maps back to the original dimension. Our belief is that increasing the dimensionality of the intermediate embedding $z$ is similar to sparse coding, which improves the generalization of the embedding, while reducing the dimensionality of the intermediate embedding $z$ serves as a bottleneck \cite{saxe2019information} to strengthen the expressiveness of the embedding $z$, as shown in following:
\begin{align} 
\label{eq:SB}
SB(z) &= M_{reduce}(Dropout(RB(z))) + z \\
\label{eq:LB}
LB(z) &= M_{scale}(Dropout(RB(z))) + z
\end{align}
where $Dropout(\cdot)$ denotes that drops out the nodes in a neural network with probability $p$, by default set to $p=0.1$; $M_{scale}(\cdot)$ and $M_{reduce}(\cdot)$ represent the expanded dimensional network and the reduced dimensional network, respectively.
\begin{equation}
\label{eq:residual-fusion}
f_{residual} = LB(SB(z))
\end{equation}
In summary, we integrate (\ref{eq:residual-block}), (\ref{eq:SB}), and (\ref{eq:LB}) to obtain the residual-based fusion, as shown in (\ref{eq:residual-fusion}). This module makes the fusion network deeper and lessens the negative effects of the deep network at the same time. 

\subsection{The Asymmetrical Contrastive Strategy}\label{sec:asymmetrical-contrastive}

\begin{figure}[htbp]
\centering
\includegraphics[width=3.3in]{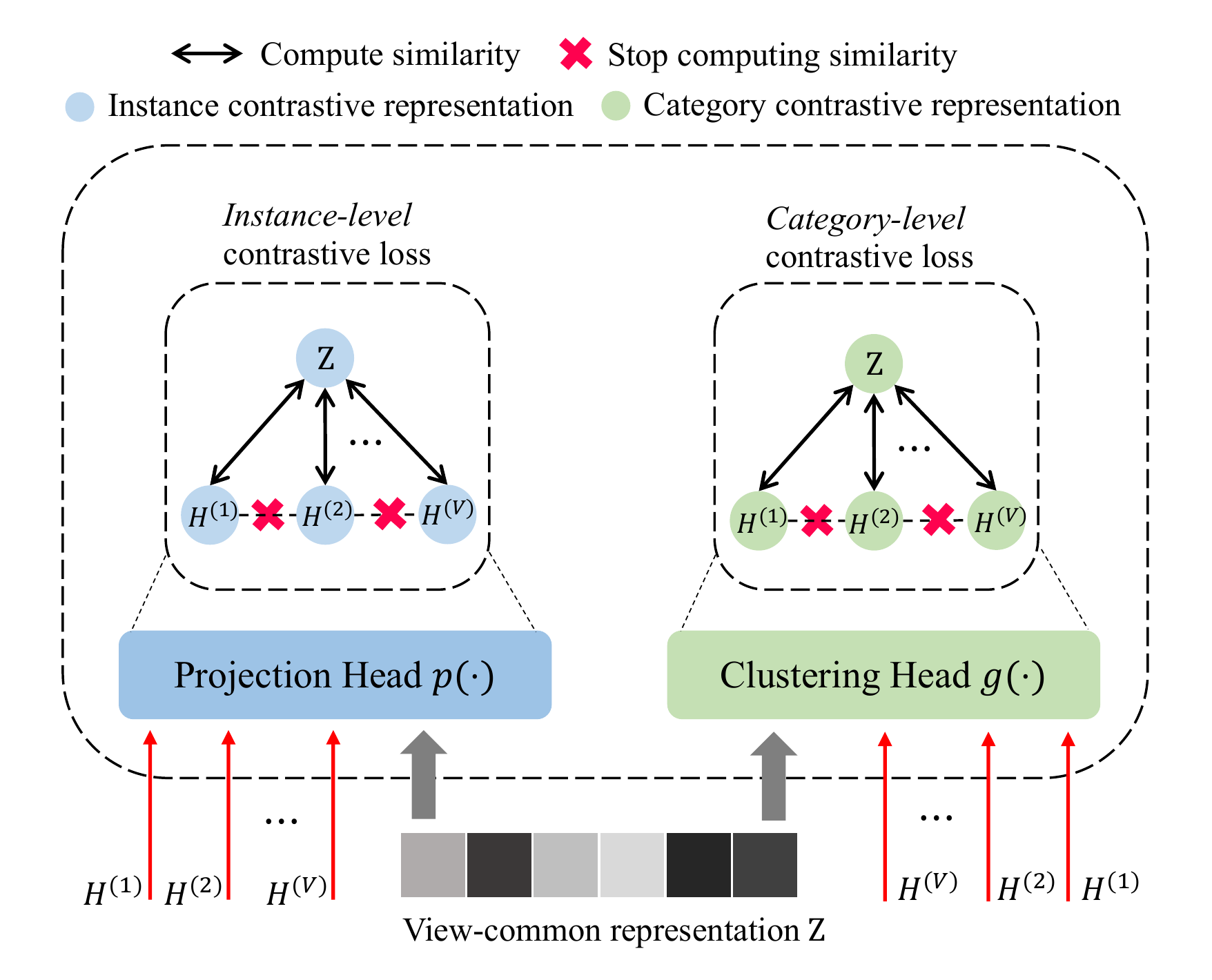}
\caption{Illustration of the workflow of the asymmetrical contrastive module}
\label{pic:acm-workflow}
\end{figure}

Contrastive learning has demonstrated great promise in the multi-view community as it seeks to maximize the mutual information between two different representations \cite{tain2020cmc, lin2021completer, trosten2021reconsidering}. However, previous works aligned the view-specific representations directly to obtain view-common representations, which has two shortcomings. Firstly, it may destroy the intrinsic structure of the views, ultimately reducing generalization. Secondly, in case of model collapse\cite{wang2020understanding}, which is when $\mathbf{Z} \equiv \{\mathbf{H}^{(i)}\}^V_{i=1} \equiv \mathbf{0}$, all view information is lost. We determine that this phenomenon is resultant of the distinction between the single-view and multi-view settings. In a single view setting, two different augmentations of one sample are used for training, while the original sample is used for testing. Even if the model collapses, the ability of the original sample remains unimpaired. In contrast, the multi-view setting presupposes that distinct views are augmentations in various forms and that the original view becomes indistinguishable after collapse. 

We hypothesize that view-specific representations, which are aligned in an additional representation space, rather than directly interacting with each other, will perform well. We refer to this alignment as an asymmetrical contrastive strategy. The asymmetrical contrastive strategy has two advantages: first, it ensures that the intrinsic structure of the view is preserved; second, the additional representations contain all the information of the view-specific representation. Consequently, the additional representation can be used directly as the view-common representation. According to \cite{chen2020simclr}, we calculate the similarity of two representations $\textbf{a}$ and $\textbf{b}$ as the cosine similarity:
\begin{equation}
\label{eq:cosine}
sim(\textbf{a}, \textbf{b}) = \frac{\textbf{a}^T \textbf{b}}{\|\textbf{a}\| \cdot \|\textbf{b}\|}.
\end{equation}
Following \cite{chen2020simclr}, we design a multi-view \textit{instance-level} contrastive loss:
\begin{equation}
\label{eq:instance-contrastive}
 \mathcal{L}_{icl} = - \sum^{V}_{v=1} \sum^{N}_{i=1}log \frac{\exp(sim(p(\mathbf{Z}_i), p(\mathbf{H}^{(v)}_i)) / \tau)}{\sum^{2N}_{j=1} \bold{1}_{[ j \neq i]} \exp(sim(p(\mathbf{Z}_i), p(\mathbf{H}^{(v)}_j)) / \tau) } 
\end{equation}
where $p(\cdot)$ is a projection head consisting of three layers of fully-connected networks used to map the representation to contrastive space\cite{chen2020simclr}; $(i, i)$ and $(i, j)$ represent a pair of positive examples and a pair of negative examples, respectively; $\bold{1}_{[ k \neq i]} \in \{ 0, 1\}$ denotes an indicator function evaluating to $1$ if $ j \neq i $, and $\tau$ is a temperature parameter with a default setting of $\tau = 0.07$ in \cite{chen2020simclr}, but we found that $\tau = 0.5$ worked well in our experiments. According to the asymmetrical contrastive strategy, we fixed the view-common representation $\mathbf{Z}$ and maximize the mutual information between $\mathbf{Z}$ and view-specific representations $\{\mathbf{H}^{(v)}\}^V_{v=1}$ in (\ref{eq:instance-contrastive}).

The instance-level contrastive loss, as shown in \cite{wang2020understanding}, aligns all samples uniformly on the hypersphere, which results in semantically similar samples lacking compactness. Drawing inspiration from the approach proposed by \cite{li2021contrastive}, we propose a multi-view \textit{category-level} contrastive loss:
\begin{equation}
\label{eq:category-contrastive}
 \mathcal{L}_{ccl} = - \sum^{V}_{v=1} \sum^{k}_{i=1}log \frac{\exp(sim(g(\mathbf{Z}_i), g(\mathbf{H}^{(v)}_i)) / \tau)}{\sum^{k}_{j=1} \bold{1}_{[ j \neq i]} \exp(sim(g(\mathbf{Z}_i), g(\mathbf{H}^{(v)}_j)) / \tau) } 
\end{equation}
where $g(\cdot)$ is a clustering head, outputting the probability of $k$ clusters. \modify{The objective function (\ref{eq:category-contrastive}) first assigns soft labels to different samples using the clustering head $g(\cdot)$, and then maximizes category consistency by forcing positive samples to be assigned to the same category.} The clustering head is built in the three-layer structure with the first two layers being fully connected networks and the last layer being the softmax function:
\begin{equation}
\label{eq:softmax}
\psi(\textbf{a})_i = \frac{e^{\textbf{a}_i}}{\sum_{j=1}^{k}e^{\textbf{a}_k}} \quad \text{for} \quad i = 1, 2, \cdots , k
\end{equation}
where $\textbf{a}_i \in \mathbb{R}^{k}$ denotes the category representation. Additionally, to avoid the suboptimal solution of assigning most instances to the same cluster, we introduce the entropy of cluster assignment probabilities \cite{hu2017learning}:
\begin{equation}
\label{eq:entropy-of-cluster}
H(\textbf{a}|\textbf{b}) \equiv -\frac{1}{k}\sum_{i=1}^{k}g(\textbf{a})_i\mathrm{log} g(\textbf{a})_i + g(\textbf{b})_i\mathrm{log} g(\textbf{b})_i
\end{equation}

In summary, we incorporate instance-level and category-level contrastive loss into an asymmetrical contrastive module:
\begin{equation}
\label{eq:entropy-of-cluster}
\mathcal{L}_{contrast} = \frac{1}{V}(\mathcal{L}_{icl} +  \mathcal{L}_{ccl} - \sum_{i=1}^{V}H(\mathbf{Z}|\mathbf{H}^{(i)})),
\end{equation}
and we present the ACM's workflow in \figurename{\ref{pic:acm-workflow}}.

\subsection{The Clustering-guided Mechanism}\label{sec:clustering-guided-mechanism}
According to our observations, the key to improving the robustness of the view-common representations is to reduce the semantic distance between the multi-view representations. Previous studies\cite{lin2021completer, ke2022efficient} incorporated contrastive learning to guide the model when learning consistent information. This approach includes cross-view reconstruction, which we argue learns both consistent and redundant information, such as image background noise and audio noise. Contrastive loss and reconstruction loss are contradictory since contrastive learning is meant to learn \textit{concepts}, while reconstruction loss requires the model to remember \textit{details}. To address this limitation, we propose a feasible guiding mechanism that is compatible with contrastive loss: clustering. To ensure simplicity and compatibility, we use an off-the-shelf online clustering approach called deep divergence-based clustering (DDC)\cite{kampffmeyer2019deep}. The DDC method uses Cauchy-Schwarz divergence (CS-divergence) to enforce cluster separability and compactness. The loss function of DDC consists of three terms, as shown below:
\begin{eqnarray}
\label{eq:ddc} \nonumber \\
\mathcal{L}_{ddc} 
=& \frac{1}{k} \sum^{k-1}_{i=1} \sum_{j>i}^{k} \frac{ \alpha_{i}^T \mathbf{Q} \alpha_{j} }{ \sqrt{\alpha_{i}^T \mathbf{Q} \alpha_{i} \alpha_{j}^T \mathbf{Q} \alpha_{j}} } \nonumber \\
+& triu(\mathbf{A}\mathbf{A}^T)  \nonumber \\
+& \frac{1}{k} \sum^{k-1}_{i=1} \sum_{j>i}^{k} \frac{ m_{i}^T \mathbf{Q} m_{j} }{ \sqrt{m{i}^T \mathbf{Q} m_{i} m_{j}^T \mathbf{Q} m_{j}}}
\end{eqnarray}
where $\mathbf{A}$ is a $n \times k$ cluster assignment matrix, $\alpha_i$ is the $i$-th column of matrix $\mathbf{A}$; $\mathbf{Q}$ denotes kernel similarity matrix computed by $\mathbf{Q}_{ij} = \exp(- \|\mathbf{z}_i - \mathbf{z}_j \|^2 / (2\sigma)^2 )$, and $\sigma$ is Gaussian kernel bandwidth, default as $0.15$; $triu(\mathbf{A}\mathbf{A}^T)$ denotes the strictly upper triangular elements of $AA^T$; $m_{i,j} = \exp(\|a_i - e_j \|^2 )$ where $e_j$ is corner $j$ of the standard simplex in $\mathbb{R}^k$. In particular, the matrix $\mathbf{A}$ consists of the output of the clustering head, i.e., $\mathbf{A} = g(\mathbf{Z})$, and $\mathbf{Q}$ is computed by the clustering head's hidden features. The first term in (14) maximizes the consistency between soft cluster assignment matrix $\mathbf{A}$ and the hidden layer matrix $\mathbf{Q}$. The second term aims to prevent all samples from being assigned to the same cluster, by requiring the assignment matrix $\mathbf{A}$ to be orthogonal. The last term enforces closeness to a corner of the simplex. In CLOVEN, to reduce the computational cost, we apply DDC to constrain the view-common representation $\mathbf{Z}$ instead of all view-specific representations.

\subsection{The Objective Function}\label{sec:the-obj-function}
To sum up, we incorporate the objective function of CLOVEN following:
\begin{equation}
\label{eq:loss}
\mathcal{L} = \mathcal{L}_{contrast} + \mathcal{L}_{ddc}.
\end{equation}
It is obvious from (\ref{eq:loss}) that the optimization of CLOVEN is a one-stage and end-to-end process. Note that in $\mathcal{L}_{ccl}$ and $\mathcal{L}_{ddc}$ share a common clustering head $g(\cdot)$. Additionally, a dynamic weight parameter might be used to balance the two losses ($\mathcal{L}_{contrast}$ and $\mathcal{L}_{ddc}$) throughout the training process\cite{groenendijk2020multi}, but in fact, we find that simply adding the two losses works well enough.

\section{Experiments}\label{sec:experiments}

\subsection{Dataset}\label{sec:dataset}
We evaluate CLOVEN and compared methods using five well-known multi-view datasets containing raw image and vector data. These are: Edge-MNIST (\textbf{E-MNIST}) ~\cite{liu2016coupled}, which is a large-scale and widely-used benchmark dataset consisting of 70,000 handwritten digit images with $28 \times 28$ pixels. The views contain the original digits and the edge-detected version, respectively; Edge-FMNIST (\textbf{E-FMNIST}) ~\cite{xiao2017fashion}, which is a more complex dataset than standard MNIST consisting of $28 \times 28$ grayscale images of clothing items. We synthesize the second view by running the same edge detector used to create Edge-MNIST; COIL-20 and COIL-100 \cite{nene1996columbia}, which depicts from different angles containing grayscale images of 20 items and RGB images of 100 items, respectively. We create a three-view dataset by randomly grouping the images for an item into three groups; Scene-15 \cite{fei2005bayesian}, which contains 4,485 images with outdoor and indoor scene environments. We use the multi-view version provided by \cite{lin2021completer}, which consists of PHOG and GIST features. 

\subsection{Baseline models and Metrics}\label{sec:baseline-models}
We compare our proposed method, CLOVEN$_{nn}$ with vanilla MLP fusion in (\ref{eq:vanilla-fusion}) and CLOVEN$_{res}$ with residual-based MLP fusion in (\ref{eq:residual-fusion}), and the following 13 baseline clustering and classification methods, which are categorized into three types:
\begin{itemize}
	\item Single-view methods: K-means (\textbf{KM}) for clustering, and Support Vector Machine (\textbf{SVM}) for classification. Note that KM$_{v=1}$ means evaluating the first view's representations, whereas KM$_{cat}$ denotes concatenating all view-specific representations.
	\item Multi-view methods: Deep Canonically Correlated Analysis (\textbf{DCCA}) \cite{andrew2013deep}, Deep Canonically Correlated Autoencoders (\textbf{DCCAE})  \cite{wang2015deep}, Autoencoder in Autoencoder Networks (\textbf{AE$^2$Nets}) \cite{zhang2019ae2}, End-to-End Adversarial-Attention Network (\textbf{EAMC}) \cite{zhou2020eamc}, Deep Adversarial Multi-view Clustering Network (\textbf{DAMC}) \cite{li2019damc}, Cross Partial Multi-View Networks (\textbf{CPMNets}) \cite{zhang2020deep}, and Efficient Multi-view Clustering Networks (\textbf{EMC-Nets}) \cite{ke2022efficient}. Note that CCA-based methods, i.e., DCCA and DCCAE, are limited to two views; for datasets with more than two views, we select the best two views based on their performance.
	\item Contrastive-based multi-view methods: Incomplete multi-view clustering via contrastive prediction (\textbf{COMPLETER})\cite{lin2021completer}, Contrastive Fusion Networks for Multi-view Clustering (\textbf{CONAN})\cite{ke2021conan}, Contrastive Multi-View Clustering (\textbf{CoMVC})\cite{trosten2021reconsidering}, and Multi-Level Feature Learning for Contrastive Multi-View Clustering (\textbf{MFLVC}) \cite{xu2022multi}.
\end{itemize}

To evaluate the performance of clustering, we apply three well-known metrics to the comparative experiments, including clustering accuracy (ACC$_{clu}$), normalized mutual information (NMI), and adjusted rand index (ARI). Given sample $x_i \in \mathbf{X}^{(v)}$ for any $i \in \{1,2,\cdots,N\}$, the predicated clustering label and the real label are indicated as $y_i$ and $c_i$, respectively. The ACC$_{clu}$ is defined as:
\begin{equation}
\label{eq:clustering-acc}
	ACC_{clu} = \frac{\sum_{i=1}^{N} \delta(y_i, map(c_i))}{N}
\end{equation}
where $y_i \in \mathbf{Y}$ represents ground-truth labels and $c_i \in \mathbf{C}$ denotes predicted clustering labels; $\delta(a, b)$ is the indicator function, i.e., $\delta(a, b) = 1$ if $a = b$, and $\delta(a, b) = 0$ otherwise; $map(\cdot)$ is the mapping function corresponding to the best one-to-one assignment of clusters to labels implemented by the Hungarian algorithm \cite{kuhn1955hungarian}; Then NMI is computed by:
\begin{equation}
\label{eq:nmi}
	NMI = \frac{I(\mathbf{Y}; \mathbf{C})}{\frac{1}{2}(H(\mathbf{Y})+H(\mathbf{C}))}
\end{equation}
$I(\cdot; \cdot)$ and $H(\cdot)$ represent mutual information and entropy functionals, respectively. ARI characterizes the agreement between two partitions $\mathbf{Y}$ and $\mathbf{C}$, defined as:  
\begin{equation}
\label{eq:ari}
	ARI = \frac{\sum_{ij}\tbinom{q_{ij}}{2} - \left[\sum_{i}\tbinom{a_{i}}{2}\sum_{j}\tbinom{b_{j}}{2}\right] / \tbinom{q}{2}}{\frac{1}{2} \left[\sum_{i}\tbinom{a_{i}}{2} + \sum_{j}\tbinom{b_{j}}{2}\right] - \left[\sum_{i}\tbinom{a_{i}}{2}\sum_{j}\tbinom{b_{j}}{2}\right]/ \tbinom{q}{2}}
\end{equation}
where $[q_{ij}] = |\mathbf{Y} \cap \mathbf{C}|$, $a_i = |\mathbf{C}|$, and $b_j = |\mathbf{Y}|$. It should be noted that the validation process of the clustering methods involves only the cases where ground truth labels $\mathbf{Y}$ are available. As for classification task, we compute classification accuracy (ACC$_{cls}$), Precision (P), and F-score to report classification results, as shown below.
\begin{equation}
\label{eq:precision}
P = \frac{TP}{TP + FP}
\end{equation}
\begin{equation}
\label{eq:fscore}
Fscore = \frac{2 \times P \times R}{P + R}
\end{equation}
where $TP$ and $FP$ are the number of true positives and the number of false positives, respectively; $R = \frac{TP}{TP+FN}$, where $FN$ is the number of false negatives. Higher values of all of the aforementioned metrics indicate better performance.

\subsection{Implementation Details}\label{sec:implementation-details}

We implement CLOVEN and other non-linear comparison methods on the PyTorch 1.10 platform, running on Ubuntu 18.04 LTS utilizing a NVIDIA GeForce RTX 3090 Graphics Processing Units (GPUs) with 24 GB memory size. We describe the implementation details in the following:

\subsubsection{Networks setting}
For fair comparisons, we employ the same view-specific encoders for all comparative methods. Specifically, we use the same convolutions networks used in CoMVC \cite{trosten2021reconsidering} for E-MNIST and E-FMNIST; We set the dimensionality of the view-specific encoders to $20-1024-1024-1024-128$ and $59-1024-1024-1024-128$ for Scene-15's view 1 and view 2, respectively; We employ the plain ResNet-18 \cite{he2016deep} for the COIL-20 and COIL-100, and all views use the same encoder to reduce computational costs. Note that we set the layers of our proposed fusion layer (the vanilla fusion and the residual-based fusion) to $L=2$ by default in all experiments except for the ablation studies.

\subsubsection{Training and testing}We use the Adam optimization technique with default parameters and an initial learning rate of $0.0001$ for both clustering and classification. For clustering, we treat the whole dataset as the training set and validation set, shuffling the training set during training. We train our model for 100 epochs and set the size of the mini-batch to 128 for E-MNIST and E-FMNIST, 64 for COIL-20 and COIL-100, and 256 for Scene-15, respectively. Specifically, we train the model 10 times with the different random seeds and report the result from the lowest value of \eqref{eq:loss}. For classification, we split the dataset into a training set and a test set in the ratio $8:2$, extracting all view-common representation to train the SVM model. The best value from the test set is reported. For training and testing the single-view method such as k-means and SVM, we use the principal component analysis (PCA) to reduce the dimensionality of each view and to be consistent with the dimensionality extracted by the deep learning method. 

\begin{table*}[ht]
\renewcommand{\arraystretch}{1.3}
\caption{Clustering results on five datasets, where “–” denotes the dataset cannot handle such scenarios. The best and the second best values are highlighted in {\color{red}red} and {\color{blue}blue}, respectively. \modify{The {\color{forestgreen}green} and {\color{gray}gray} arrows represent the improvement rate and decline rate, respectively, relative to the second-best method. All results of the comparison methods were obtained by our reproduced version.}}
\label{tab:clustering-result}       
\begin{center}
\scalebox{0.87}{\begin{tabular}{cccccccccccccccc}
\hline\noalign{\smallskip}
 & \multicolumn{3}{c}{E-MNIST} & \multicolumn{3}{c}{E-FMNIST} & \multicolumn{3}{c}{COIL-20} & \multicolumn{3}{c}{COIL-100} & \multicolumn{3}{c}{Scene-15} \\
\noalign{\smallskip} Method & ACC$_{clu}$ & NMI & ARI & ACC$_{clu}$ & NMI & ARI & ACC$_{clu}$ & NMI & ARI & ACC$_{clu}$ & NMI & ARI & ACC$_{clu}$ & NMI & ARI \\
\noalign{\smallskip}\hline\noalign{\smallskip}
KM$_{v=1}$ & 47.54 & 41.54 & 29.86 & 49.06 & 50.18 & 34.45 & 53.54 & 64.17 & 37.12 & 41.29 & 68.06 & 33.69 & 36.68 & 36.94 & 18.99 \\
KM$_{v=2}$ & 38.54 & 36.39 & 21.65 & 52.33 & 51.44 & 35.29 & 55.83 & 65.02 & 39.85 & 44.17 & 66.58 & 31.82 & 29.05 & 27.53 & 12.39 \\
KM$_{v=3}$ & - & - & - & - & - & - & 55.83 & 65.40 & 41.28 & 39.42 & 65.34 & 26.74 & - & - & - \\
KM$_{cat}$ & 48.27 & 42.87 & 30.07 & 51.82 & 51.71 & 34.36 & 56.25 & 70.38 & 42.28 & 45.17 & 70.64 & 34.81 & 29.68 & 28.73 & 13.24 \\
\noalign{\smallskip}\hline\noalign{\smallskip}
DCCA (2013) \cite{andrew2013deep} & 36.48 & 30.01 & 18.43 & 35.47 & 29.19 & 16.61 & 42.29 & 51.09 & 28.79 & 31.88 & 42.66 & 22.79 & 31.1 & 33.39 & 15.21 \\
DCCAE (2015) \cite{wang2015deep} & 39.19 & 33.75 & 22.3 & 39.34 & 41.24 & 25.96 & 71.88 & 80.48 & 61.73 & 34.58 & 63.19 & 23.53 & 35.88 & 35.26 & 17.37 \\
Ae$^2$-Nets (2019) \cite{zhang2019ae2}  & 43.57 & 43.88 & 40.68 & 44.96 & 45.07 & 38.55 & 56.04 & 73.96 & 50.8 & 45.71 & 71.73 & 37.32 & 28.7 & 31.46 & 13.8 \\
DAMC (2019) \cite{li2019damc} & 63.27 & 59.06 & 51.82 & 50.16 & 55.38 & 43.72 & 62.39 & 68.51 & 59.1 & 36.67 & 44.72 & 30.27 & 36.16 & 40.6 & \textcolor{blue}{25.42} \\
EAMC (2020) \cite{zhou2020eamc} & 66.13 & 63.55 & 61.86 & 54.52 & \textcolor{red}{63.25} & 43.88 & 69.14 & 73.75 & 64.3 & 37.84 & 58.11 & 29.65 & 37.28 & 40.48 & 24.31 \\
EMC-Nets (2022) \cite{ke2022efficient} & 69.31 & 62.29 & 62.43 & 53.15 & 55.38 & 40.1 & 71.55 & 76.41 & 70.24 & 41.82 & 68.63 & 28.11 & 31.06 & 29.87 & 16.97 \\
\noalign{\smallskip}\hline\noalign{\smallskip}
COMPLETER (2021) \cite{lin2021completer} & 84.51 & 75.48 & 68.84 & 54.58 & 54.36 & \textcolor{red}{51.63} & 77.62 & 86.48 & 72.73 & 45.46 & 77.8 & 42.77 & \textcolor{blue}{39.20} & \textcolor{blue}{42.81} & 23.95 \\
CONAN (2021) \cite{ke2021conan} & 80.51 & 76.13 & 69.2 & 57.25 & 55.27 & 42.05 & 72.53 & 80.39 & 70.68 & 50.67 & 77.4 & 40.09 & 36.77 & 37.33 & 20.33 \\
CoMVC (2021) \cite{trosten2021reconsidering} & \textcolor{red}{91.71} & \textcolor{blue}{82.67} & \textcolor{blue}{80.67} & \textcolor{blue}{59.03} & 49.38 & 40.55 & \textcolor{blue}{89.96} & \textcolor{blue}{90.53} & \textcolor{blue}{74.26} & 48.0 & 80.61 & 44.24 & 38.71 & 40.78 & 22.39 \\
MFLVC (2022) \cite{xu2022multi} & 81.92 & 75.11 & 65.31 & 57.27 & 55.85 & 31.23 & 73.12 & 82.9 & 65.94 & \textcolor{blue}{52.29} & \textcolor{blue}{83.21} & \textcolor{red}{49.43} & 31.66 & 34.65 & 14.89 \\
CLOVEN$_{nn}$ (Ours) & 82.13 & 78.25 & 75.14 & 58.33 & 55.19 & 40.86 & 73.58 & 84.72 & 71.95 & 51.52 & 79.38 & 41.27 & 37.54 & 38.29 & 20.88 \\
CLOVEN$_{res}$ (Ours) & \textcolor{blue}{91.28} & \textcolor{red}{83.06} & \textcolor{red}{82.1} & \textcolor{red}{61.01} & \textcolor{blue}{57.61} & \textcolor{blue}{44.14} & \textcolor{red}{91.46} & \textcolor{red}{91.13} & \textcolor{red}{82.16} & \textcolor{red}{54.83} & \textcolor{red}{85.31} & \textcolor{blue}{48.84} & \textcolor{red}{43.21} & \textcolor{red}{42.82} & \textcolor{red}{25.62} \\
\noalign{\smallskip}\hline\noalign{\smallskip}
$\Delta$ SOTA  & \textcolor{gray}{$\downarrow$0.43} & \textcolor{forestgreen}{$\uparrow$0.39} & \textcolor{forestgreen}{$\uparrow$1.43} & \textcolor{forestgreen}{$\uparrow$1.98} & \textcolor{gray}{$\downarrow$5.64} & \textcolor{gray}{$\downarrow$7.49} & \textcolor{forestgreen}{$\uparrow$1.5} & \textcolor{forestgreen}{$\uparrow$0.6} & \textcolor{forestgreen}{$\uparrow$7.9} & \textcolor{forestgreen}{$\uparrow$2.54} & \textcolor{forestgreen}{$\uparrow$2.1} & \textcolor{gray}{$\downarrow$0.59} & \textcolor{forestgreen}{$\uparrow$1.98} & \textcolor{forestgreen}{$\uparrow$0.01} & \textcolor{forestgreen}{$\uparrow$0.2} \\
\noalign{\smallskip}\hline
\end{tabular}}
\end{center}
\end{table*}

\begin{table*}[ht]
\renewcommand{\arraystretch}{1.3}
\caption{Classification results on five datasets. where “–” denotes the dataset cannot handle such scenarios. The best and the second best values are highlighted in {\color{red}red} and {\color{blue}blue}, respectively. \modify{The {\color{forestgreen}green} and {\color{gray}gray} arrows represent the improvement rate and decline rate, respectively, relative to the second-best method. All results of the comparison methods were obtained by our reproduced version.}}
\label{tab:classification-result}       
\begin{center}
\scalebox{0.83}{\begin{tabular}{cccccccccccccccc}
\hline\noalign{\smallskip}
 & \multicolumn{3}{c}{E-MNIST} & \multicolumn{3}{c}{E-FMNIST} & \multicolumn{3}{c}{COIL-20} & \multicolumn{3}{c}{COIL-100} & \multicolumn{3}{c}{Scene-15} \\
\noalign{\smallskip} Method & ACC$_{cls}$ & P & F-score & ACC$_{cls}$ & P & F-score & ACC$_{cls}$ & P & F-score & ACC$_{cls}$ & P & F-score & ACC$_{cls}$ & P & F-score \\
\noalign{\smallskip}\hline\noalign{\smallskip}
SVM$_{v=1}$ & 39.58 & 40.32 & 39.09 & 73.31 & 74.75 & 73.42 & 9.44 & 5.96 & 6.11 & 20.83 & 19.27 & 16.81 & 48.83 & 45.65 & 45.74 \\
SVM$_{v=2}$ & 33.92 & 34.10 & 32.50 & 66.66 & 66.72 & 66.41 & 17.71 & 16.30 & 13.42 & 7.12 & 7.60 & 5.65 & 49.61 & 49.09 & 46.40 \\
SVM$_{v=3}$ & - & - & - & - & - & - & 25.00 & 29.49 & 25.19 & 6.52 & 4.31 & 4.62 & - & - & - \\
SVM$_{cat}$ & 42.89 & 41.08 & 41.11 & 73.51 & 74.83 & 73.58 & 10.42 & 9.64 & 8.02 & 10.00 & 7.81 & 6.99 & 57.41 & 57.38 & 53.99 \\
\noalign{\smallskip}\hline\noalign{\smallskip}
DCCA (2013) \cite{andrew2013deep} & 73.65 & 73.64 & 73.32 & \textcolor{blue}{78.1} & \textcolor{blue}{77.52} & \textcolor{blue}{77.48} & 47.92 & 52.37 & 46.43 & 33.83 & 33.81 & 33.61 & 55.63 & 52.29 & 52.49 \\
DCCAE (2015) \cite{wang2015deep} & 90.44 & 90.34 & 90.35 & 74.78 & 74.51 & 74.23 & 89.62 & 88.26 & 88.12 & 30.42 & 20.39 & 21.66 & 54.96 & 51.82 & 51.24 \\
Ae$^2$-Nets (2019) \cite{zhang2019ae2}  & 92.41 & 92.39 & 92.37 & 74.26 & 74.09 & 74.02 & 78.12 & 74.88 & 72.32 & 58.54 & 57.58 & 54.9 & 57.41 & 58.03 & 55.06 \\
CPMNets (2020) \cite{zhang2020deep}  & 92.73 & 92.18 & 92.55 & 77.42 & 75.69 & 75.43 & \textcolor{blue}{90.11} & \textcolor{blue}{89.52} & \textcolor{blue}{89.64} & 60.79 & 60.17 & 60.32 & 55.47 & 55.33 & 55.92 \\
\noalign{\smallskip}\hline\noalign{\smallskip}
COMPLETER (2021) \cite{lin2021completer} & 92.51 & 92.12 & 91.95 & 73.43 & 74.29 & 69.99 & 89.58 & 86.96 & 86.13 & 55.83 & 51.41 & 50.18 & \textcolor{red}{70.18} & \textcolor{red}{70.47} & \textcolor{red}{70.05} \\
CONAN (2021) \cite{ke2021conan} & 93.67 & 93.51 & 93.22 & 74.51 & 74.16 & 74.38 & 87.32 & 85.11 & 82.17 & 54.72 & 50.39 & 51.6 & 41.15 & 40.31 & 37.10 \\
CLOVEN$_{nn}$ (Ours) & \textcolor{blue}{94.76} & \textcolor{blue}{93.98} & \textcolor{blue}{94.14} & 77.88 & 77.24 & 77.16 & 88.43 & 87.28 & 85.69 & \textcolor{blue}{80.6} & \textcolor{blue}{79.51} & \textcolor{blue}{78.64} & 58.22 & 58.22 & 58.23 \\
CLOVEN$_{res}$ (Ours) & \textcolor{red}{96.23} & \textcolor{red}{96.25} & \textcolor{red}{96.24} & \textcolor{red}{80.47} & \textcolor{red}{79.86} & \textcolor{red}{79.88} & \textcolor{red}{93.75} & \textcolor{red}{94.25} & \textcolor{red}{94.52} & \textcolor{red}{99.79} & \textcolor{red}{99.8} & \textcolor{red}{99.75} & \textcolor{blue}{63.43} & \textcolor{blue}{61.43} & \textcolor{blue}{61.51} \\
\noalign{\smallskip}\hline\noalign{\smallskip}
$\Delta$ SOTA  & \textcolor{forestgreen}{$\uparrow$1.47} & \textcolor{forestgreen}{$\uparrow$2.27} & \textcolor{forestgreen}{$\uparrow$2.1} & \textcolor{forestgreen}{$\uparrow$2.37} & \textcolor{forestgreen}{$\uparrow$2.34} & \textcolor{forestgreen}{$\uparrow$2.4} & \textcolor{forestgreen}{$\uparrow$3.64} & \textcolor{forestgreen}{$\uparrow$4.73} & \textcolor{forestgreen}{$\uparrow$4.88} & \textcolor{forestgreen}{$\uparrow$19.19} & \textcolor{forestgreen}{$\uparrow$19.2} & \textcolor{forestgreen}{$\uparrow$20.29} & \textcolor{gray}{$\downarrow$6.75} & \textcolor{gray}{$\downarrow$9.04} & \textcolor{gray}{$\downarrow$8.54} \\
\noalign{\smallskip}\hline
\end{tabular}}
\end{center}
\end{table*}

\subsection{Experimental Results Analysis}\label{sec:results-analysis}

In this section, we conducted experimental comparisons among clustering and classification methods. We tested single-view methods, generative model-based multi-view methods, and contrastive learning-based multi-view methods on five datasets. In terms of clustering performance, we evaluated ACC$_{clu}$, NMI, and ARI. In terms of classification performance, we evaluated ACC$_{cls}$, Precision (P), and F-score. For single-view methods, we validated the performance of each view representation, as well as the concatenated view representation. We present the results of clustering and classification in Table \ref{tab:clustering-result} and Table \ref{tab:classification-result}, respectively. We also measured the classification running time of each method to measure the effectiveness of view-common representation, and report the results in Table \ref{tab:runningtime-result}.

\subsubsection{\modify{Comparison with single-view methods}} Our proposed method, CLOVEN, outperforms the best single-view algorithms in terms of clustering and classification on five datasets. For example, on the E-MNIST dataset, CLOVEN achieves 43.01\%, 40.19\%, and 52.03\% higher ACC$_{clu}$, NMI, and ARI scores respectively, compared to the best results of K-means. Moreover, CLOVEN also achieves 53.34\%, 55.17\%, and 55.13\% higher ACC$_{cls}$, Percision, and F-score, respectively, compared to the best results of SVM. In addition, by analyzing the classification performance of SVM on the COIL-20/100 dataset, we found that simply concatenating all view-specific representations does not directly improve the classification performance. This observation confirms our hypothesis that shallow fusion is not applicable to semantically complex scenarios.

\subsubsection{\modify{Comparison with multi-view methods}} Our proposed method outperforms or closely matches the performance of the best multi-view algorithm on most datasets. Interestingly, the single-view method outperforms some multi-view methods, such as DCCA, DCCAE and Ae$^2$-Nets, on datasets such as E-MNIST and F-MNIST. This suggests that shallow fusion techniques like concatenation or weighted-sum may not always be the best approach for integrating multiple views. 

In contrast to generative model-based multi-view methods like DAMC, EAMC, COMPLETER, and MFLVC, CLOVEN achieves superior performance on high-dimensional datasets like COIL-20 and COIL-100. We argue that higher-dimensional data requires the network storing more detailed information for reconstruction, causing a shift in the balance between the view consistency and complementarity. Some approaches, like EAMC, COMPLETER, and MFLVC, utilize additional constraints like GAN or contrastive learning methods to narrow the performance gap. Our position is that unconstrained use of contrastive learning on view-specific representations for information alignment does not effectively balance view consistency and complementarity. It is worth noting that CoMVC and CLOVEN achieved similar results on the E-MNIST dataset. CoMVC uses contrastive learning to identify information-rich views before fusing data from multiple views with a weighted-sum approach. In contrast, our fusion strategy assumes equal information for each view and then balances the fused representations to achieve a higher-order semantic alignment. We argue that the former is suitable for simple data fusion scenarios, while the latter is more appropriate for complex structure fusion scenarios.

Our comparison with COMPLETER reveals that our method outperforms it on all datasets except for Scene-15. We hypothesize that the Scene-15 dataset is not a low-level dataset (extracted by PHOG and GIST), and that the generative model memorizes specific details to enhance classification performance. In contrast, on low-level datasets, such as COIL-20/100, COMPLETER memorizes a large amount of view-specific information due to the presence of a large amount of redundant information (can be seem as noise) in the images, but its classification performance is diminished. Therefore, when considering the clustering results as well, we can conclude that our method is superior for extracting conceptual information.

\subsubsection{\modify{Evaluation of the representation efficiency}} Table \ref{tab:runningtime-result} illustrates that CLOVEN outperforms all other methods in terms of average time spent on the classification task. As the dimensionality of the view-common representation increases, the computational cost of downstream tasks also increases, even with the same number of datasets. Furthermore, our method stands out for its efficiency, taking only 71.38 seconds to complete the classification task on E-MNIST, in contrast to the simple fusion methods which require twice as much time as CLOVEN. This result emphasizes the importance of optimization in fusion strategies for downstream tasks and highlights the remarkable benefits of the view-common representation that our method creates, significantly facilitating the downstream task.

Based on the clustering and classification results, our findings conclude that: i) our proposed deep fusion method has the ability to extract a more expressive and task-friendly view-common representation for the downstream task when compared to the shallow fusion strategy. ii) In addition, compared to other contrastive learning-based methods, our designed asymmetrical contrastive strategy achieves a better balance between view-consistent and -complementary information.

\begin{table}[htbp]
\renewcommand{\arraystretch}{1.3}
\caption{The Classification running time of results on five datasets, where \textit{Dim.} indicates the dimensionality of view-common representation, \textit{Time} is the duration in seconds required to perform the SVM. The best and the second best values are highlighted in {\color{red}red} and {\color{blue}blue}, respectively. \modify{The {\color{forestgreen}green} and {\color{gray}gray} arrows represent the improvement rate and decline rate, respectively, relative to the second-best method. All results of the comparison methods were obtained by our reproduced version.}}
\label{tab:runningtime-result}       
\begin{center}
\scalebox{0.70}{\begin{tabular}{ccccccccccc}
\hline\noalign{\smallskip}
 & \multicolumn{2}{c}{E-MNIST} & \multicolumn{2}{c}{E-FMNIST} & \multicolumn{2}{c}{COIL-20} & \multicolumn{2}{c}{COIL-100} & \multicolumn{2}{c}{Scene-15} \\
\noalign{\smallskip} Method & Dim. & Time & Dim. & Time  & Dim. & Time & Dim. & Time & Dim. & Time  \\
\noalign{\smallskip}\hline\noalign{\smallskip}
SVM$_{cat}$ & 576 & 416.67 & 576 & 458.89 & 1024 & \textcolor{blue}{0.04} & 1024 & 3.40 & 256 & \textcolor{red}{0.67}  \\
DCCA & 288 & 160.47 & 288 & 410.0 & 512 & 0.05 & 512 & 1.08 & 128 & 0.87  \\
DCCAE & 288 & 162.21 & 288 & 458.39 & 512 & 0.06 & 512 & 1.05 & 128 & 0.89  \\
Ae$^2$Nets & 576 & 413.15 & 576 & 633.0 & 1024 & 0.06 & 1024 & 1.14 & 256 & 1.48  \\
CPMNets & 576 & 487.62 & 576 & 620.31 & 1024 & \textcolor{blue}{0.04} & 1024 & 1.28 & 256 & 1.62  \\
COMPLETER & 576 & 896.24 & 576 & 823.38 & 1024 & 0.05 & 1024 & 1.39 & 256 & 1.55  \\
CONAN & 288 & \textcolor{blue}{43.79} & 288 & \textcolor{blue}{355.16} & 512 & \textcolor{blue}{0.04} & 512 & \textcolor{blue}{0.92} & 128 & 1.35  \\
Ours & 288 & \textcolor{red}{71.38} & 288 & \textcolor{red}{319.22} & 512 & \textcolor{red}{0.03} & 512 & \textcolor{red}{0.68} & 128 & \textcolor{blue}{0.77}  \\
\noalign{\smallskip}\hline\noalign{\smallskip}
$\Delta$ SOTA  & - & \textcolor{forestgreen}{$\downarrow$72.41} & - & \textcolor{forestgreen}{$\downarrow$35.94} & - & \textcolor{forestgreen}{$\downarrow$0.01} & - & \textcolor{forestgreen}{$\downarrow$0.24} & - & \textcolor{gray}{$\uparrow$0.1}  \\
\noalign{\smallskip}\hline
\end{tabular}}
\end{center}
\end{table}

\begin{figure*}[t]
\centering
\includegraphics[width=7in]{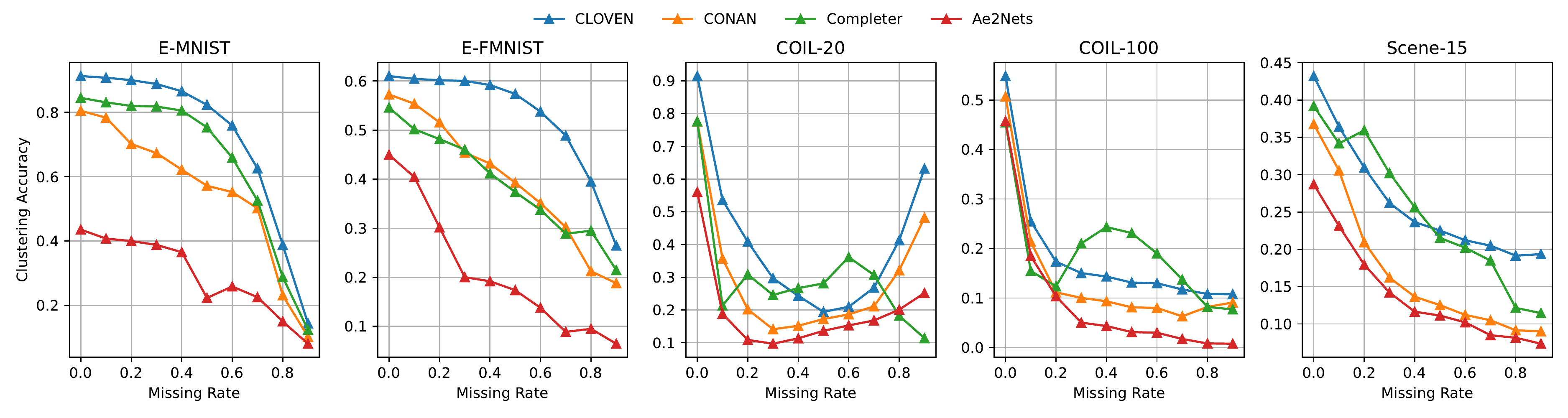}
\caption{Illustration of the change in clustering accuracy on five datasets from the TCTI scenario. }
\label{pic:tcti-results}
\end{figure*}

\begin{figure*}[htbp]
\centering
\includegraphics[width=7in]{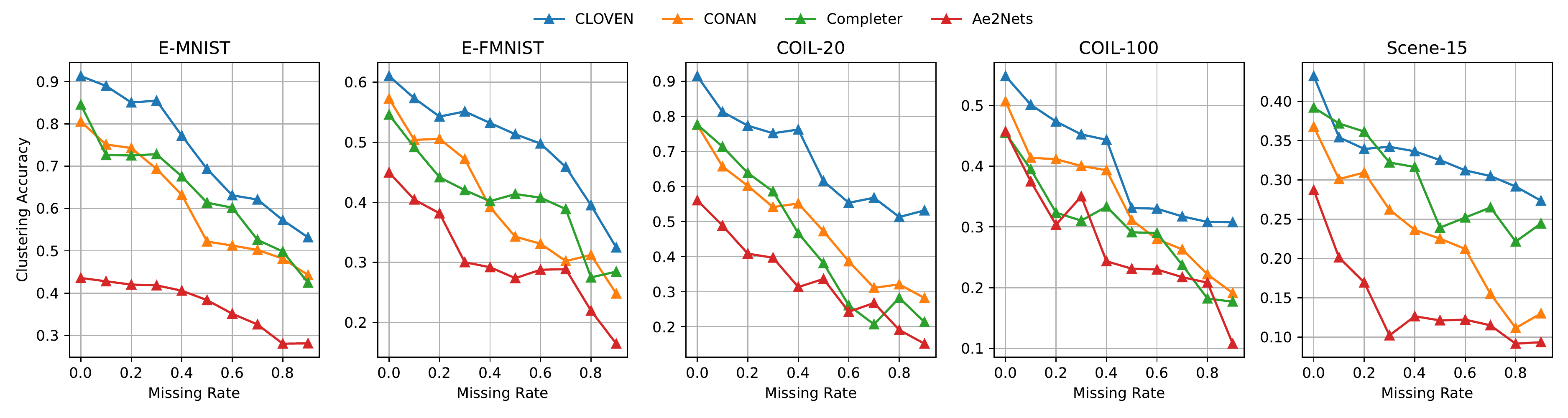}
\caption{Illustration of the change in clustering accuracy on five datasets from the TITI scenario.}
\label{pic:titi-results}
\end{figure*}

\begin{figure}[htbp]
\centering
\includegraphics[width=3.5in]{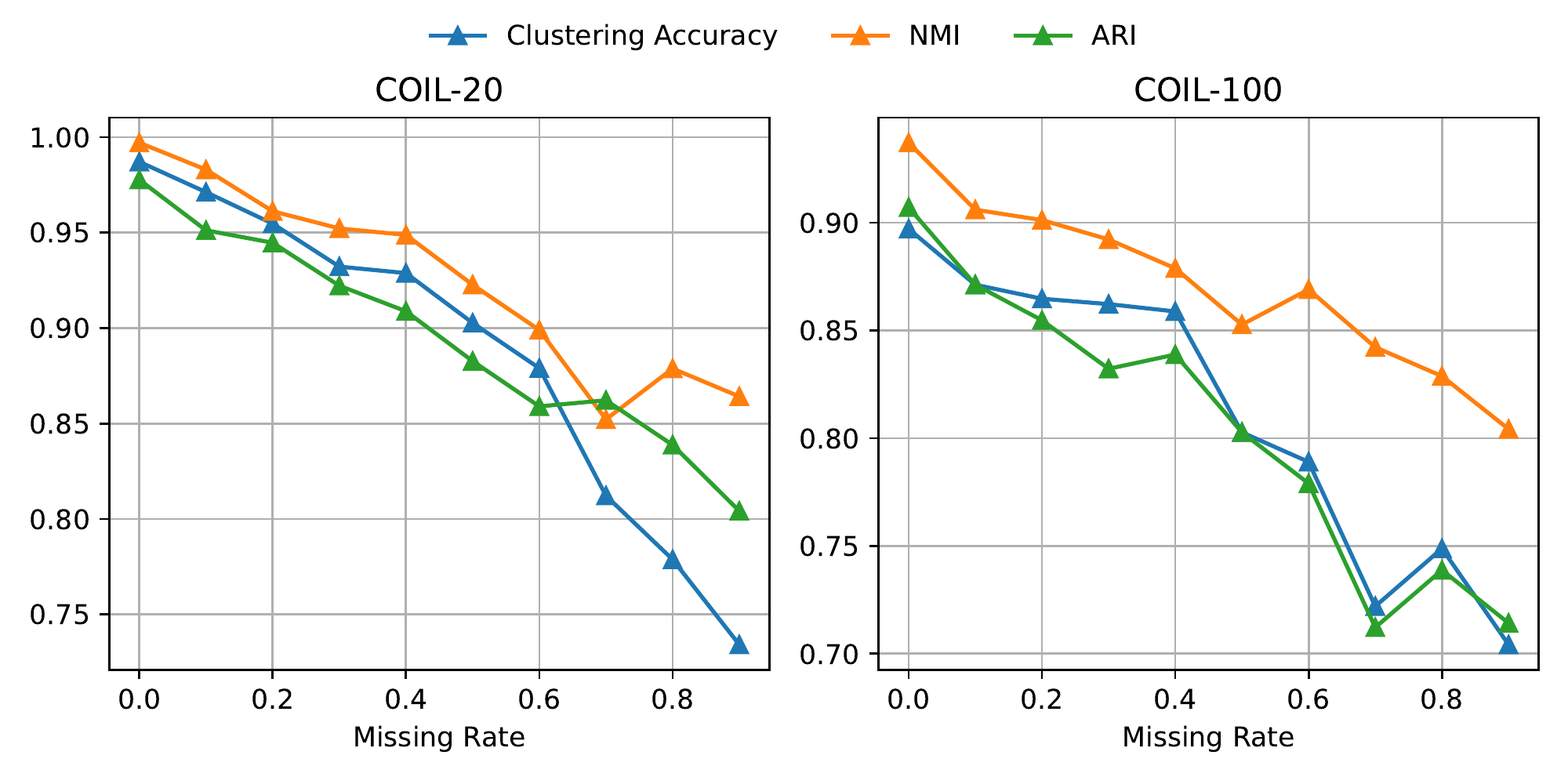}
\caption{Illustration of the CLOVEN's change in clustering accuracy, NMI, and ARI on the COIL-20 and COIL-100 datasets using a view-specific encoder pre-trained on ImageNet-1K.}
\label{pic:pretrain-results}
\end{figure}

\subsection{Experiments on Incomplete View Setting}\label{sec:incomplete}
We designed two scenarios, Training with Complete Testing with Incomplete (TCTI) and Training with Incomplete Testing with Incomplete (TITI), to assess the robustness of CLOVEN under incomplete views. In the TCTI scenario, our model demonstrated superior robustness compared to other comparison methods for both the E-MNIST and E-FMNIST datasets. Even when the missing rate ranged from 0.4 to 0.7, CLOVEN's clustering accuracy exceeded that of other methods. These results indicate that our method effectively mitigates the effects of noise in various incomplete scenarios.

In addition, we observed that most algorithms in the TCTI scenario show a pattern of clustering accuracy decreasing and increasing as the missing rate on the COIL-20 dataset increases. This is because COIL-20 has less data, making it more vulnerable to under-fitting and noise attacks compared to the powerful view-specific encoder used by the algorithms. On the other hand, the network in the TITI scenario is less likely to be disrupted by noise because it can directly learn and reduce the noise through asymmetrical alignment. To test this hypothesis, we conducted an experiment using pre-trained ResNet-18 as the view-specific encoder and found that under-fitting was eliminated. This suggests that the view-specific encoder's capability is a critical factor in determining the model's robustness. See \figurename{\ref{pic:pretrain-results}} for the results of this experiment.

\begin{figure*}[htbp]
\centering
\includegraphics[width=6.9in]{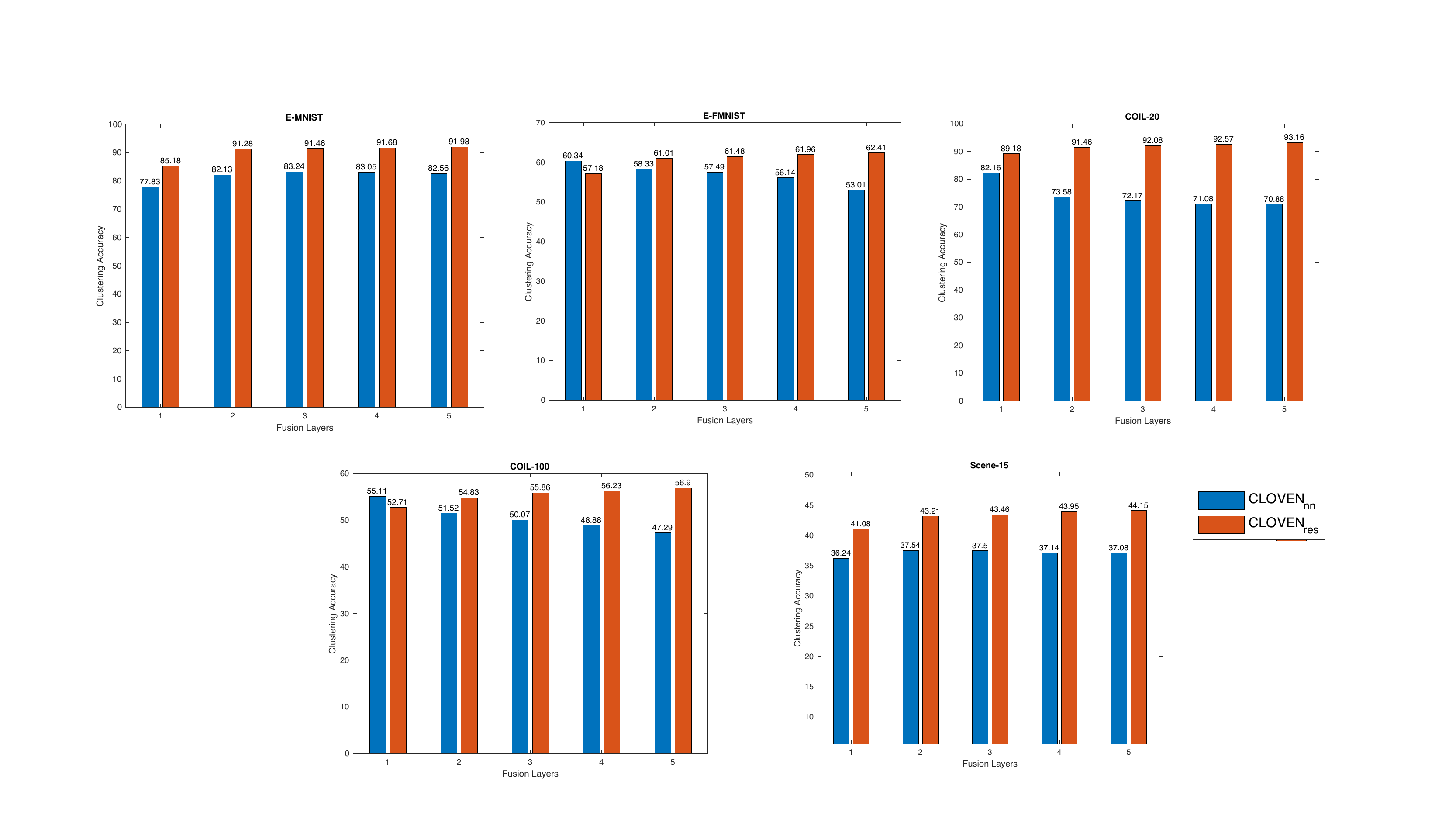}
\caption{Illustration of the effect of fusion layers on five datasets.}
\label{pic:fusion-layers}
\end{figure*}

\begin{table}[htbp]
\renewcommand{\arraystretch}{1.3}
\caption{The contributions of each component on the E-FMNIST dataset.}
\label{tab:ab_module}       
\begin{center}
\scalebox{0.68}{\begin{tabular}{ccccccccccc}
\hline\noalign{\smallskip}
Case & $\mathcal{L}_{icl}$ & $\mathcal{L}_{ccl}$ & $\mathcal{L}_{ddc}$  & Asym. & ACC$_{clu}$ & NMI & ARI & ACC$_{cls}$ & P & F-score\\
\noalign{\smallskip}\hline\noalign{\smallskip} 
Baseline & $\checkmark$ & $\checkmark$ & $\checkmark$ & $\checkmark$ & 61.01 & 57.61 & 44.14 & 80.47 & 79.86 & 79.88 \\
\noalign{\smallskip}\hline\noalign{\smallskip} 
& $\checkmark$ &  &  & $\checkmark$ & 32.64 & 30.87 & 28.61 & 72.15 & 71.38 & 71.03 \\
W/o cluste-& & $\checkmark$ &  & $\checkmark$ & 28.55 & 26.11 & 24.36 & 70.52 & 70.19 & 70.18 \\
ring-guided & $\checkmark$ & $\checkmark$ &  & & 52.61 & 48.8 & 35.69 & 73.28 & 73.07 & 73.12 \\
 & $\checkmark$ & $\checkmark$ &  & $\checkmark$ & 54.87 & 50.10 & 39.25 & 76.39 & 76.14 & 76.19 \\
\noalign{\smallskip}\hline\noalign{\smallskip} 
W/o &  &  &  $\checkmark$ &  & 45.75 & 44.51 & 40.30 & 69.88 & 69.26 & 69.53 \\
asym. & $\checkmark$ & $\checkmark$ & $\checkmark$ &  & 58.83 & 55.37 & 42.64 & 77.19 & 77.10 & 77.10 \\
\noalign{\smallskip}\hline
\end{tabular}}
\end{center}
\end{table}

\begin{figure*}[htbp]
\centering
\includegraphics[width=7in]{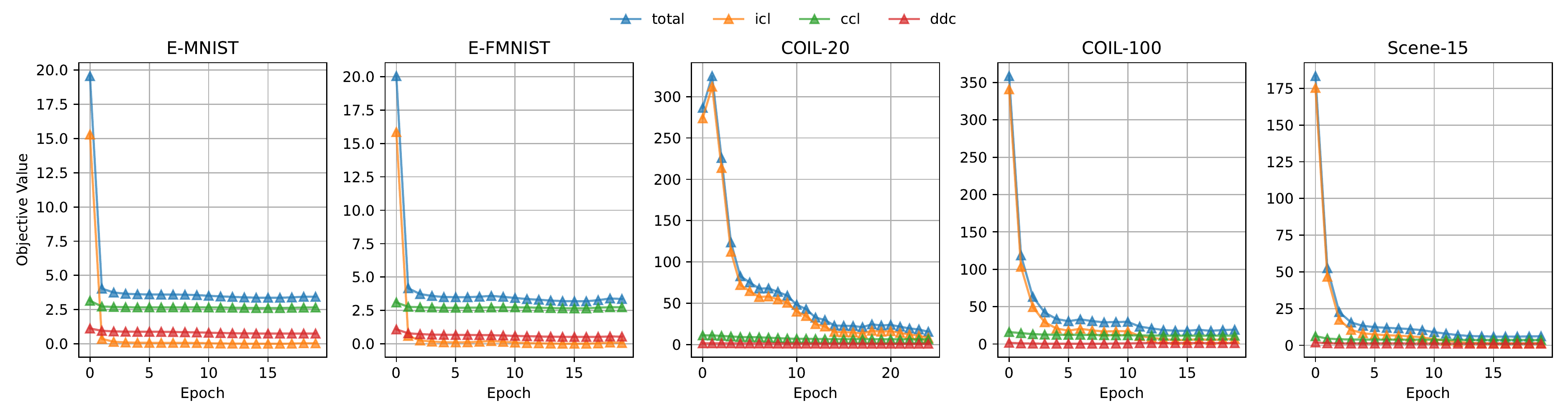}
\caption{Convergence analysis on the proposed CLOVEN using all the multi-view datasets.}
\label{pic:convergence}
\end{figure*}

\begin{figure*}[htbp]
\centering
\includegraphics[width=7in]{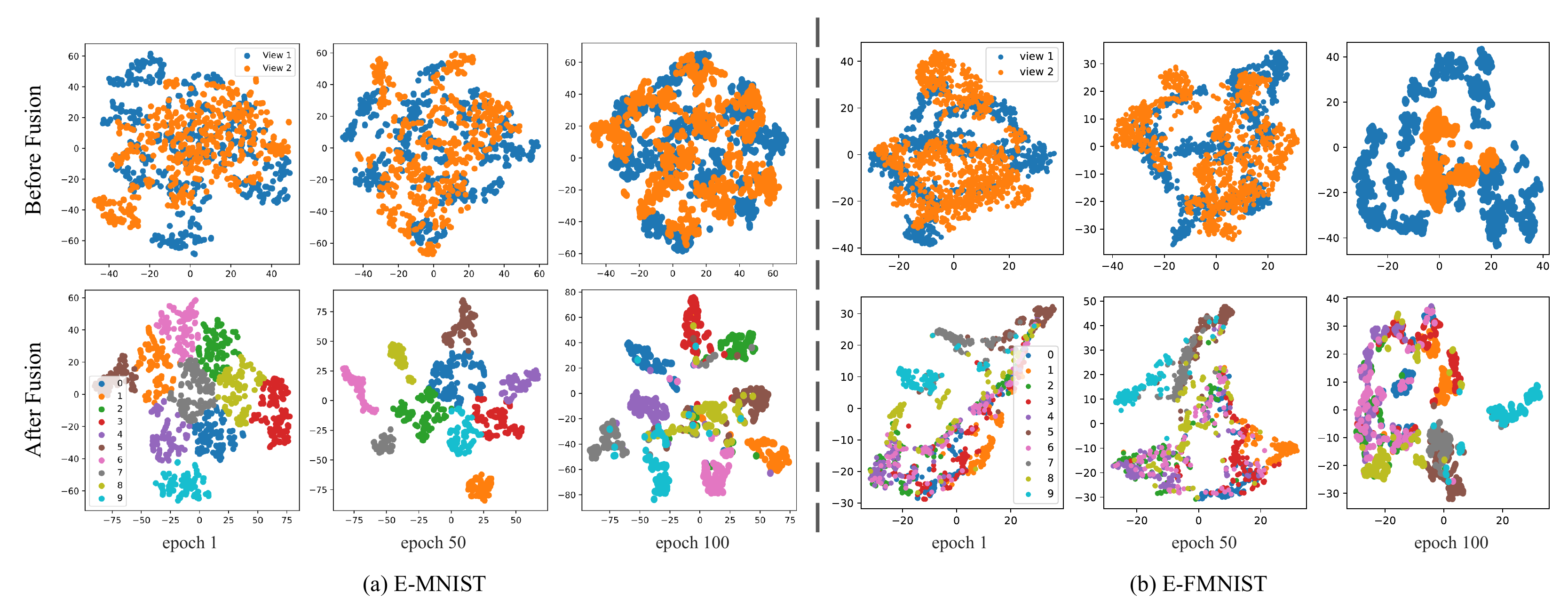}
\caption{Visualization of representations on the E-MNIST (left) and E-FMNIST (right) dataset before (first row) and after fusion (second row) using T-SNE \cite{van2008visualizing} at epoch 1st, 50th, and 100th.}
\label{pic:visualization}
\end{figure*}

\subsection{\modify{Ablation Studies}}\label{sec:ablation-study}

\subsubsection{The effect of fusion layers}
We evaluate the clustering performance of CLOVEN$_{nn}$ and CLOVEN$_{res}$ with varying numbers of fusion network layers ($L$) on different datasets, as presented in Fig. \ref{pic:fusion-layers}. The results demonstrate that increasing the number of fusion network layers could improve model's performance on most of the datasets. The most noticeable enhancement in performance occurs when the network layers increase from 1 to 2. Even though increasing the layers results in a slight improvement, we consider $L = 2$ to be the most effective way due to the fact that network complexity will increase with the increased layers. However, for the COIL-20 and COIL-100 datasets, we found that the clustering accuracy of the vanilla fusion network decreases as the number of layers increases. In the absence of any optimization tricks, such as skip-connections, the vanilla MLPs are probably more prone to obtaining suboptimal solutions than CLOVEN$_{res}$ networks, which incorporate residual connections.

\subsubsection{Is the asymmetric contrastive fusion necessary} We conducted two groups of ablation studies to evaluate the necessity of the asymmetric contrastive strategy. The results are shown in Table \ref{tab:ab_module}. Note that the first row of Table \ref{tab:ab_module} is the complete version of CLOVEN. The results of the first group are presented in the first and seventh rows of Table \ref{tab:ab_module}, and the results of the second group are presented in the fourth and fifth rows of Table \ref{tab:ab_module}. Both sets of data indicate that the model using the asymmetric contrastive strategy performs better than the model without this module. The experimental results confirmed our hypothesis: the asymmetric contrastive strategy helps to retain the complementarity of views and indirectly improves the generalization of view-common representations.

\subsubsection{Is clustering-guided mechanism necessary} According to the results in the second, third, and sixth rows of Table \ref{tab:ab_module}, it can be concluded that the clustering-guided mechanism has a greater improvement on the clustering and classification performance than using instance-level or category-level versions alone. Comparison of the results in the fifth and seventh rows shows that using he clustering-guided mechanism is more effective in improving the clustering and classification performance than the asymmetric contrastive fusion strategy. Moreover, the overall performance of CLOVEN is the best when the two modules are used together.


\subsubsection{Convergence analysis}
In this part, we discuss the convergence property of CLOVEN. We depict the objective values with increasing training epochs in \figurename{\ref{pic:convergence}}. As observed, the objective value of CLOVEN decreases continuously with each epoch and eventually reaches a constant value. Our model requires approximately 25 training epochs to reach convergence on the five datasets considered. This is substantially better than COMPLETER and EMC-Nets, which are autoencoder-based methods that require 500 and 1000 training epochs, respectively.

\subsection{Visualization}\label{sec:visualization}
In this section, we present the changes in representations before and after contrastive fusion on the E-MNIST and E-FMNIST datasets in \figurename{ \ref{pic:visualization}}. We visualized the changes in 2D using T-SNE \cite{van2008visualizing} for 2,000 randomly selected samples to simplify the image. The visualization demonstrates that the view-specific representations $\bold{H}^{(1)}$ and $\bold{H}^{(2)}$ retain their intrinsic structure during training. However, the view-common representation $\bold{Z}$ becomes more separable and compact with the increasing number of training epochs. Therefore, we conclude that the proposed contrastive fusion method enhances the quality of the view-common representation without compromising the view-specific representations.

\section{Conclusion}\label{sec:conclusion}
This paper introduces CLOVEN, a novel deep fusion module and an asymmetrical contrastive strategy for learning robust multi-view representations. CLOVEN offers several advantages over existing multi-view representation methods. Firstly, it achieves a better balance between view consistency and complementarity, outperforming generative models while also being faster to train. Secondly, CLOVEN preserves the view-specific intrinsic structure and achieves superior performance compared to contrastive learning-based methods, thanks to the deep fusion module and the asymmetrical contrastive strategy. Furthermore, the view-common representation extracted by CLOVEN alleviates the computational burden of downstream tasks. Experimental evaluations in clustering, classification, and incomplete view scenarios demonstrate the effectiveness and superiority of CLOVEN over 13 state-of-the-art multimodal representation learning methods. Moreover, the view-common representations extracted by CLOVEN effectively reduce computational costs for classification. It is important to note that while the efficacy of CLOVEN is supported by extensive experiments, the deep fusion module and the asymmetrical contrastive strategy lack a theoretical foundation. Therefore, we anticipate that this study will spark research interest in multi-view representation learning based on contrastive learning. In future studies, we aim to continue exploring relevant theories and evaluating the effectiveness of our approach on larger datasets.

\section*{Acknowledgments}
This work was supported by the Fundamental Research Funds for the Central Universities, China (No. 2021JBZD006 and No. 2023YJS113), and the Fundamental Research Funds for the Beijing Jiaotong University, China (No. 2021JBWZB002).

%


%

\bibliographystyle{IEEEtran}
\bibliography{paper}

%
%
%
%
%
%
%
%
%

\end{document}